\documentclass[letterpaper]{article} 
\usepackage{aaai24}  
\usepackage{times}  
\usepackage{helvet}  
\usepackage{courier}  
\usepackage[hyphens]{url}  
\usepackage{graphicx} 
\usepackage{natbib}  
\usepackage{caption} 
\usepackage{algorithm}
\usepackage{algorithmic}

\usepackage{newfloat}
\usepackage{listings}
\DeclareCaptionStyle{ruled}{labelfont=normalfont,labelsep=colon,strut=off} 
\floatstyle{ruled}
\newfloat{listing}{tb}{lst}{}
\floatname{listing}{Listing}
\usepackage{color}
\usepackage{subfigure}
\usepackage{multirow}
\usepackage{amssymb}
\usepackage{amsmath}
\usepackage{booktabs}
\usepackage{mathrsfs}
\usepackage{mathtools}

\def\G{\mathcal{G}}
\def\L{\mathcal{L}}

\def\D{\mathcal{D}}

\title{Teacher as a Lenient Expert: Teacher-Agnostic Data-Free Knowledge Distillation}

\author{
    Hyunjune Shin, Dong-Wan Choi\textsuperscript{\rm}\thanks{Corresponding Author}\\
}
\affiliations{
    \textsuperscript{\rm } Department of Computer Science and Engineering, Inha University, South Korea\\

    heounjunee@gmail.com, dchoi@inha.ac.kr\\
}

\begin{document}
\maketitle

\begin{abstract}
Data-free knowledge distillation (DFKD) aims to distill pretrained knowledge to a student model with the help of a generator without using original data. In such data-free scenarios, achieving stable performance of DFKD is essential due to the unavailability of validation data. Unfortunately, this paper has discovered that existing DFKD methods are quite sensitive to different teacher models, occasionally showing catastrophic failures of distillation, even when using well-trained teacher models. Our observation is that the generator in DFKD is not always guaranteed to produce precise yet diverse samples using the existing representative strategy of minimizing both \textit{class-prior} and \textit{adversarial} losses. Through our empirical study, we focus on the fact that class-prior not only decreases the diversity of generated samples, but also cannot completely address the problem of generating unexpectedly low-quality samples depending on teacher models. In this paper, we propose the \textit{teacher-agnostic data-free knowledge distillation} (TA-DFKD) method, with the goal of more robust and stable performance regardless of teacher models. Our basic idea is to assign the teacher model a \textit{lenient} expert role for evaluating samples, rather than a strict supervisor that enforces its class-prior on the generator. Specifically, we design a sample selection approach that takes only clean samples verified by the teacher model without imposing restrictions on the power of generating diverse samples. Through extensive experiments, we show that our method successfully achieves both robustness and training stability across various teacher models, while outperforming the existing DFKD methods.
\end{abstract}

\section{Introduction}

Knowledge distillation (KD)~\cite{KD} is a powerful compression technique that transfers the knowledge of a pretrained teacher model to a smaller student model. Typically, KD methods require data samples that are used to train the teacher model, in order to properly guide the training of the student model. However, in real-world scenarios, it is neither always possible nor desirable to assume the availability of training data. To address such practical issues, data-free knowledge distillation (DFKD) has been actively studied~\cite{KEGNET,WITH_KEGENT}, aiming to distill pretrained knowledge through the assistance of a generator, without the use of original data samples. The generator is also trained based on the teacher model to generate synthetic samples, which are intended to be replacements of the original samples in the distillation process.

\begin{figure}[t!]
    \centering
    \hspace{-7mm}\includegraphics[width= 0.95\columnwidth]{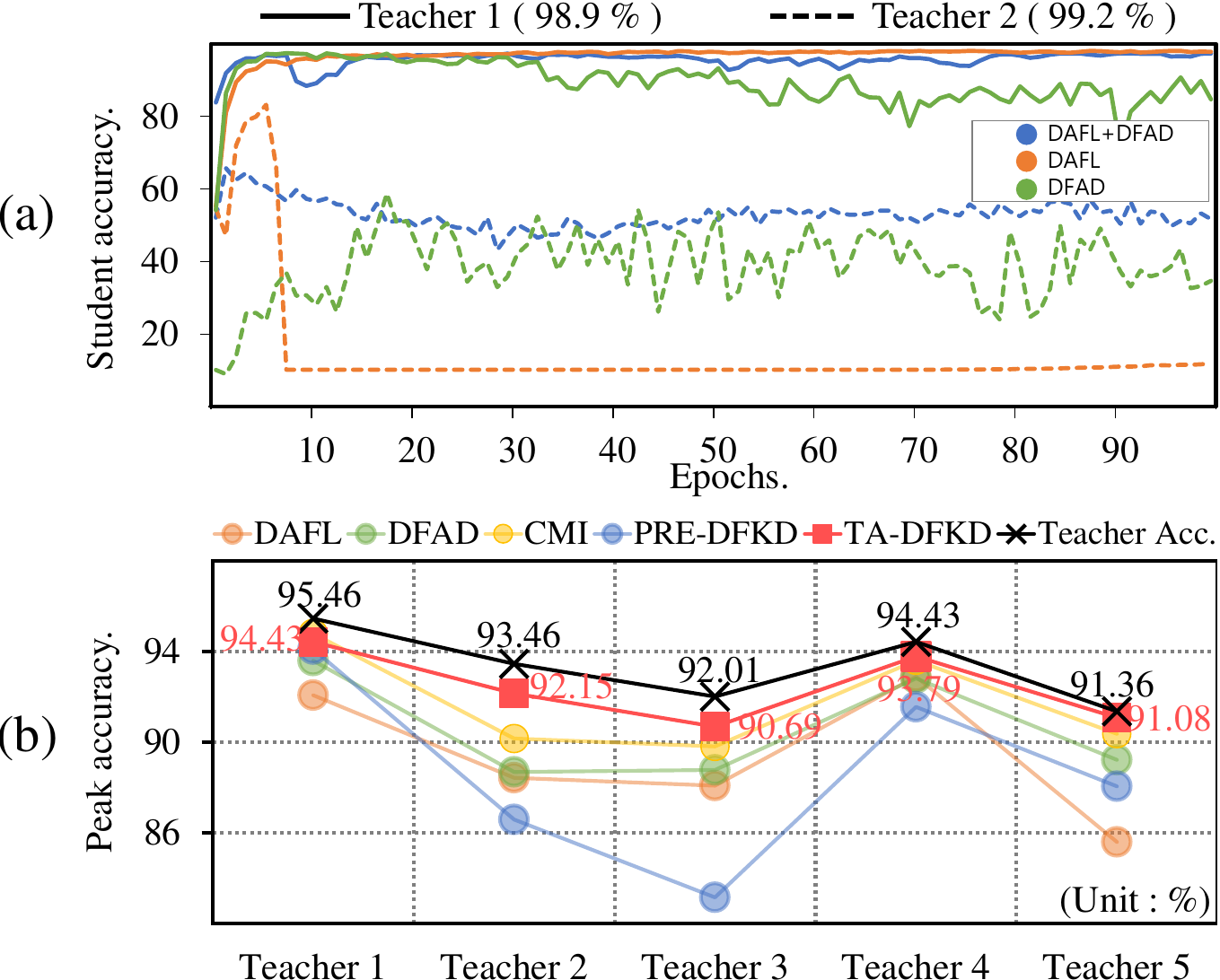}
    \caption{(a) Training curves of student models during distillation of two different well-trained teacher models on MNIST when using one or both of class-prior (DAFL) and adversarial learning (DFAD) losses. (b) Peak accuracies of student models distilled from five different teacher models on CIFAR10 when using different DFKD methods including our TA-DFKD method}
    \label{fig:problem_trim}
\end{figure}

A key challenge in DFKD arises from the unavailability of validation data, making it impossible to accurately evaluate the effectiveness of distillation. Therefore, it is crucial for DFKD methods to ensure the stable and robust performance no matter which teacher models are distilled. To this end, the state-of-the-art (SOTA) DFKD methods incorporate the following three components into their training loss function for the generator: \textit{class-prior}, \textit{adversarial}, and \textit{representation} losses. Class-prior, initially introduced by DAFL~\cite{DAFL019}, aims to generate accurate samples that can be classified by the teacher model into a specific class. On the other hand, the adversarial loss, first proposed by DFAD~\cite{DFAD} and ZSKT~\cite{DFAD_BEFORE}, intends to generate hard samples that maximize the output discrepancy between the teacher and student models, thereby enhancing the diversity of generated samples. Lastly, the representation loss focuses on learning feature-level information of real data with respect to the teacher model.

Unfortunately, this paper has discovered that existing DFKD methods are quite sensitive to different teacher models, occasionally showing catastrophic failures of distillation, even when using well-trained teacher models with high performance. As shown in Figure~\ref{fig:problem_trim}(a), we first focus on how two representative techniques, such as enforcing class-prior (e.g., DAFL) and minimizing adversarial loss (e.g., DFAD), can sometimes fail in distillation from two different teacher models on MNIST with the same level of accuracy achieved by the same training method. Although \textit{Teacher~2} is slightly better in accuracy than \textit{Teacher~1}, both DAFL and DFAD completely fail to distill the knowledge of \textit{Teacher~2}, resulting in a large performance gap from their successful counterparts using \textit{Teacher~1}. Although more and more recent proposals combine both class-prior and adversarial losses, such a mixed approach may not be a successful solution either, as shown by the failure of DAFL~+~DFAD in distilling \textit{Teacher~2} in Figure~\ref{fig:problem_trim}(a). Note that there is no difference in training strategies between \textit{Teacher~1} and \textit{Teacher~2}.

In our findings, this teacher-sensitive failure in DFKD occurs mainly due to a misguided generator that does not always produce precise yet diverse samples when employing the above two strategies, namely minimizing class-prior and adversarial losses. First of all, class-prior such as in DAFL is intended to improve the sample quality, but it also tends to guide the generator to focus on only easy samples. As a result, the student model can learn only a small fraction of the teacher's knowledge. In extreme cases, the resulting student model can misclassify every sample into a particular class, which is why the model distilled from \textit{Teacher 2} by DAFL keeps 0.1 accuracy (i.e., out of 10 digits) in Figure~\ref{fig:problem_trim}(a). On the other hand, the adversarial loss for the generator used in DAFD is effective to generate harder samples, which are possibly more diverse as well, but can lead to unrealistic samples that are not relevant to any of the classes of the teacher model. In order to achieve both high quality and diversity of synthetic samples, the recent works like \cite{CMI21,Dreming_to_distill,PREDFKD22,MB_DFKD,CU_DKFD} combine both techniques. However, depending on teacher models, we find that they are not guaranteed to find a sweet spot between two conflicting losses, one for precision and the other for diversity, and consequently suffer from the generation of unexpectedly low-quality samples. As demonstrated in Figure~\ref{fig:problem_trim}(b), none of the SOTA DFKD methods show a satisfactory level of robustness across 5 different pretrained models on CIFAR-10.

In this paper, we revisit the necessity of class-prior, which has been believed crucial by most SOTA methods, and focus on its drawback, namely enforcing the teacher's strict restriction to the generator. In our analysis, we find that a generator can freely generate more diverse samples when it is trained without class-prior. Moreover, despite the attempts of class-prior to enhance the sample quality, our observation reveals that relying only on the class-prior loss still allows the generator to produce low-quality samples, even without the adversarial loss.

Based on these observations, we propose the \textit{teacher-agnostic data-free knowledge distillation} (TA-DFKD) method that assigns the teacher model a \textit{lenient} expert role, namely removing the class-prior restriction for the generator to explore larger area in the sample space for achieving higher diversity. At the same time, in pursuit of high precision of synthetic samples, TA-DFKD utilizes the teacher model as an expert who can evaluate the quality of synthetic samples, thereby discards unexpectedly low-quality samples. Inspired by the existing works \cite{LNL_survey} on learning from noisy labels, we design a sample selection method that takes only generated samples whose labels are confirmed to be sufficiently precise by the teacher model, using the Gaussian Mixture Model. 

As observed in Figure~\ref{fig:problem_trim}(b), our TA-DFKD method demonstrates a highest level of teacher-agnostic robustness by consistently achieving the best accuracy close to those of teacher models. This trend is also observed in our extensive experimental results, where TA-DFKD manages to achieve both the robustness across various teacher models and stability at converging time of the distillation process, outperforming the existing DFKD methods.

\section{Related Works}
\subsubsection{Data-Free Knowledge Distillation}

\begin{table}[t!]
    \small
    \centering
    {
        \begin{tabular}{c|cccccc}
        \toprule
         & \textbf{\scriptsize DAFL} & \textbf{\scriptsize DFAD} & \textbf{\scriptsize ADI} & \textbf{\scriptsize CMI} & \textbf{\scriptsize PRE-DFKD}  & \textbf{\scriptsize TA-DFKD} \\ 
        \midrule
        {\textit{Cls.}} & $\checkmark$ & & $\checkmark$ & $\checkmark$ & $\checkmark$ &  \\
        \midrule
        {\textit{Adv.}} & & $\checkmark$ & $\checkmark$ & $\checkmark$ & $\checkmark$  & $\checkmark$\\
        \midrule
        \multirow{2}{*}{\textit{Rep.}} & $\checkmark$ & & $\checkmark$& $\checkmark$ & $\checkmark$ & $\checkmark$\\
        & \textit{\scriptsize activ.} & & \textit{\scriptsize BNS} & \textit{\scriptsize BNS} & \textit{\scriptsize activ.} & \textit{\scriptsize BNS}\\
        \bottomrule
        \end{tabular}
    }
    \caption{Summary of the existing DFKD methods, DAFL~\cite{DAFL019}, DFAD~\cite{DFAD}, ADI~\cite{Dreming_to_distill}, CMI~\cite{CMI21}, PRE-DFKD~\cite{PREDFKD22} and TA-DFKD (ours), in terms of using three major components, class-prior, adversarial, and representation losses.}
    \label{tab:dfkd method table}
\end{table}

In data-free knowledge distillation (DFKD), given only a pretrained teacher model without any real or meta data, our focus is on how to generate synthetic samples that can be used to effectively transfer the teacher's knowledge to a target student model. There are two initial strategies to this end, optimizing random noisy images themselves~\cite{WITH_KEGENT} or employing a generator extracted from a pretrained model~\cite{KEGNET,DAFL019}. Since the former is more computationally expensive~\cite{WITH_KEGENT,Dreming_to_distill}, recent studies have primarily focused on the latter approach, where the main issue is to train the generator only using the teacher model. Except for KegNet \cite{KEGNET}, most DFKD methods \cite{DAFL019,DFAD,DFAD_BEFORE,MB_DFKD,PREDFKD22,MAD,CU_DKFD} adopt a one-phase distillation scheme such that the generator and the student are simultaneously trained from scratch, while freezing the teacher network. This enables a progressive transfer of the teacher's knowledge using the generator being trained. To generate more effective samples, three types of loss terms are mainly leveraged for training the generator: \textit{class-prior}, \textit{adversarial}, and \textit{representation} losses, as described in the previous section. Table~\ref{tab:dfkd method table} provides a summary of which losses are employed in existing DFKD methods.

DAFL~\cite{DAFL019} first exploits class-prior that enforces the generator to produce samples that are precise enough to be well predicted by the teacher. It also proposes a representation loss, referred to as \textit{activation}, which aims to maximize activation values of the feature maps. Another representation loss, introduced in ADI~\cite{Dreming_to_distill} and called BNS, constrains the statistics of batch normalization layers stored in the teacher model. DFAD~\cite{DFAD} and ZSKT~\cite{DFAD_BEFORE} adopt an adversarial learning strategy inspired by GAN~\cite{GAN}, aiming to generate more challenging samples that maximize disagreement between the teacher and student models. This approach encourages the student model to learn diverse knowledge from the teacher model, but may lead to unrealistic samples that belongs to none of teacher's categories.

Consequently, recent studies have attempted to combine class-prior, adversarial and representation losses, aiming of generating precise and diverse samples, while proposing their additional techniques to further enhance performance. CMI~\cite{CMI21} suggests using contrastive learning to increase the diversity of generated samples. CuDFKD~\cite{CU_DKFD} and AdaDFQ~\cite{adadfq} propose adaptive learning so that the student model can progressively learn the teacher's knowledge, and ABD~\cite{ABD} deals with a scenario with untrustworthy teacher models. Furthermore, MB/PRE-DFKD~\cite{MB_DFKD,PREDFKD22} pay attention to undesirable forgetting in the student model caused by adversarial learning, as seen in the training curve of DFAD with \textit{Teacher 1} in Figure~\ref{fig:problem_trim}(a). To prevent this forgetting phenomenon, MB/PRE-DFKD~\cite{MB_DFKD,PREDFKD22} propose the use of a memory bank or an extra generative model. With the same goal, MAD~\cite{MAD} suggests employing exponential moving average for generator updates, while META-DFKD~\cite{PRE_META} incorporates meta-learning into the generator training process. Despite some synergy effects observed in these DFKD methods that combine the three losses, none of them achieve a satisfactory level of robustness and stability across different teacher models, as revealed in our experimental results.

\subsubsection{Learning from Noisy Labels}
Unlike popular benchmark datasets assuming always correct labels in deep neural networks (DNNs), data labeling in practice can be highly prone to errors, leading to noisy labels. To address this issue, there has been a branch of works, called learning from noisy labels (LNL)~\cite{LNL_survey}, which focuses on preventing a DNN from overfitting to data with noisy labels. A representative approach is sample selection that identifies clean samples by modeling the difference between clean ones and those with noisy labels. A simple policy can be taking samples with smaller loss values. More advanced strategies include using a pretrained model \cite{MENTORNET} and training dual models \cite{DECOUPLING,COTEACHING,COTEACHING2,DIVIDEMIX} to make a better decision on clean samples. Our method is inspired by these sample selection strategies in LNL even though the DFKD problem itself is not directly related to identifying noisy labels. Specifically, we employ the Gaussian Mixture Model introduced by DivideMix~\cite{DIVIDEMIX}, as it aligns well with our objective of selecting high-quality samples with respect to the pretrained teacher model.

\section{Methodology}
\subsection{Framework of Generator-Based DFKD}
In the standard generator-based DFKD framework, we consider the following three networks: a pretrained teacher model~$\theta_T$, a student model~$\theta_S$, and a generator~$\theta_G$. The ultimate goal of DFKD is the same as in the normal KD, that is, transferring the knowledge of the teacher model to the student model. Instead of real data, however, DFKD uses the generator to generate a fake sample~$\hat{x} =\theta_G(z)$ with some random vector~$z \sim p_z(z)$, and feeds these synthetic samples to $\theta_T$ and $\theta_S$ for minimizing the following distillation loss:
\begin{equation} \label{eqnkd}
\L_{KD} = \mathbb{E}_{z \sim p_z(z)}[\D(\theta_T(\theta_G(z)),~\theta_S(\theta_G(z)))],
\end{equation}
where $\D(\cdot, \cdot)$ is the distance between the outputs of two models and $p_z(z)$ is usually $\mathcal{N}(0, 1)$. The most challenging issue here is how to define an effective loss function~$\L_G$ to train $\theta_G$ without using any real data. To this end, most DFKD methods employ a mixed loss function as follows:
\begin{equation} \label{eqn:lg:prev}
\L_{G} = \alpha\L_{Cls} + \beta\L_{Adv} + \gamma\L_{Rep},
\end{equation}
where $\L_{Cls}$ is the class-prior loss, $\L_{Adv}$ is the adversarial loss, and $\L_{Rep}$ is the representation loss. 
Given $\L_{KD}$ and $\L_{G}$, while freezing $\theta_T$, the final goal of DFKD is to simultaneously train $\theta_S$ and $\theta_G$ with the following objective functions: $\min\limits_{\theta_S} \L_{KD}$ and $\min\limits_{\theta_G} \L_{G}$.

\subsubsection{Our Findings.}
In this work, we argue that the generator is not always guaranteed to synthesize precise yet diverse samples for various teacher models, despite minimizing $\mathcal{L}_{G}$ in Eq. (\ref{eqn:lg:prev}). In particular, we focus on the catastrophic failure of DAFL \cite{DAFL019} in Figure~\ref{fig:problem_trim}(a), which heavily relies on class-prior and thus reveals the drawbacks of class-prior when training the generator, namely decreasing sample diversity yet allowing low-quality samples.

\subsection{Revisiting Class-Prior in DFKD}
With the goal of generating more accurate samples, the class-prior loss $\L_{Cls}$ is usually defined as:
$$
\L_{Cls} = \mathbb{E}_{z \sim p_z(z)}[\ell_{ce}(\theta_T(\theta_G(z)),~\hat{y}_z)],
$$
where $\hat{y}_z$ is a one-hot vector corresponding to the class with the maximum probability in $\theta_T(\theta_G(z))$ and $\ell_{ce}(\cdot, \cdot)$ is the cross-entropy loss function. Since $\L_{Cls}$ will continue to incur loss values with some extent until $\theta_T(\theta_G(z))$ becomes close to the one-hot vector, the generator~$\theta_G$ will be more and more focused on producing less challenging samples, rather than exploring various sample cases that might be useful for transferring the teacher's knowledge. This will potentially reduce the overall diversity of generated samples, leading to less effective distillation from the teacher. Based on our intuition, this subsection conducts a detailed experimental analysis on class-prior, considering its necessity in the generator loss function.

\subsubsection{Lower Sample Diversity.}

\begin{figure}[t!]
    \hspace{-7mm}
    \centering
    \includegraphics[width=0.7\columnwidth]{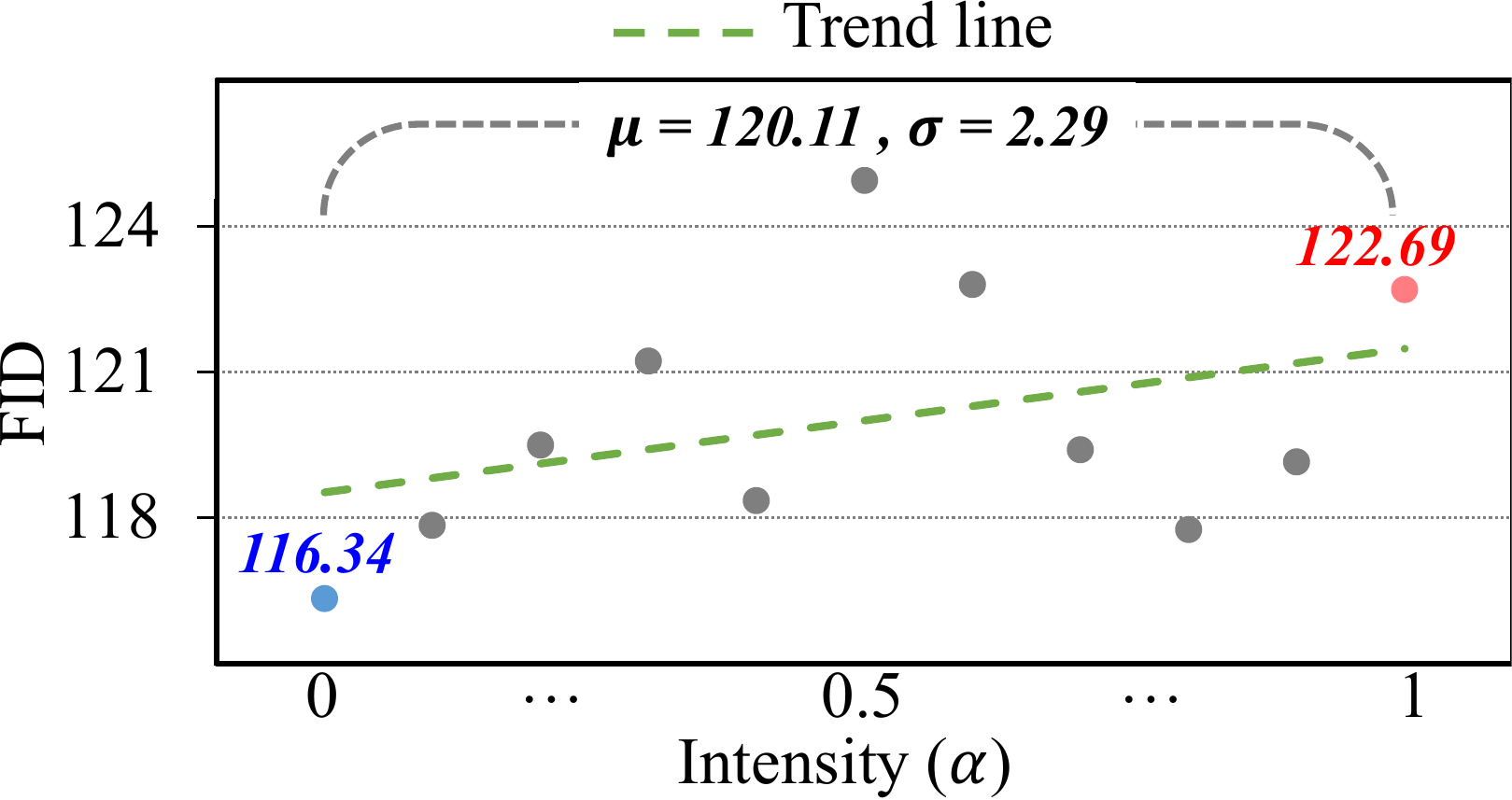}
    \caption{FID scores using a pretrained ResNet-34 model on CIFAR-10 with class-prior's intensity values from 0 to 1.}
    \label{fig:FID_official}    
\end{figure}

\begin{figure}[t!]
    \centering
        
    \subfigure[\label{fig:modecollapse_pic}With class-prior]{\includegraphics[width=0.45\columnwidth]{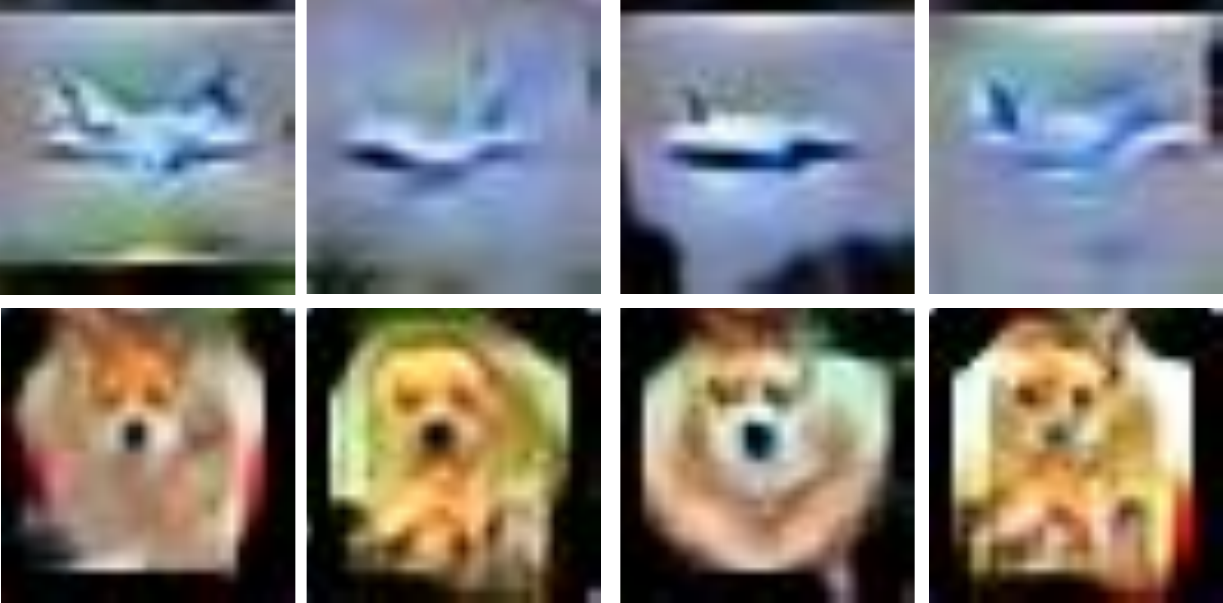}}
    \hspace{2mm}\subfigure[\label{fig:non modecollapse_pic}Without class-prior]{\includegraphics[width=0.45\columnwidth]{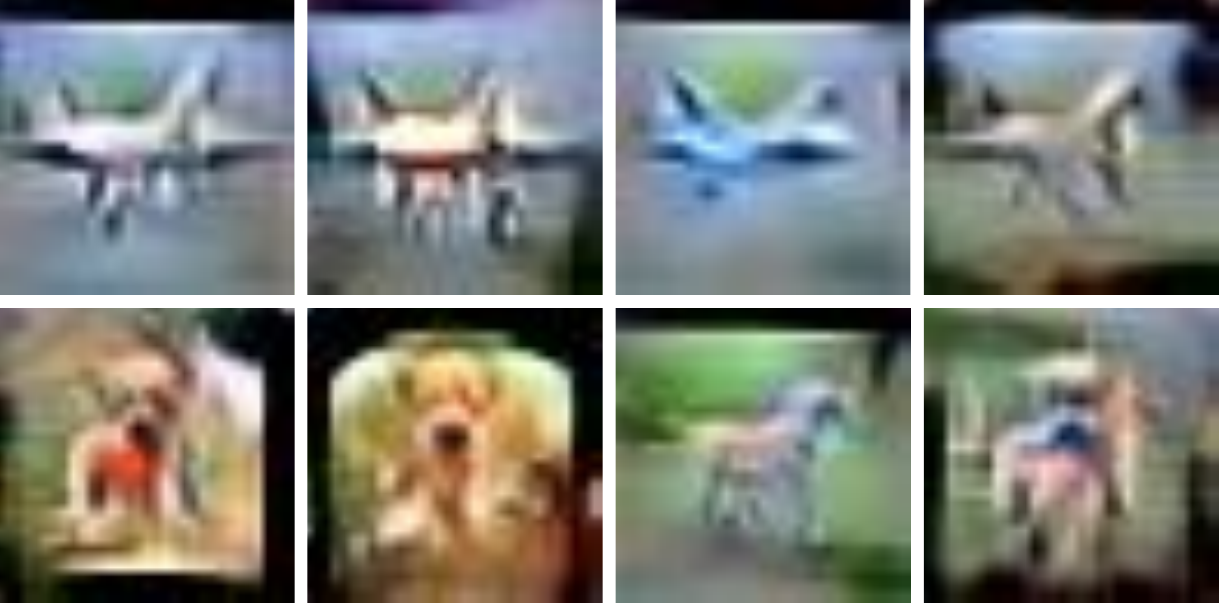}}
    
    \caption{Images of \textit{Airplane} (top) and \textit{Dog} (bottom) generated by each trained version of the generator with or without class-prior for a pretrained ResNet-34 model on CIFAR-10.}
    \label{fig: mode collapse}

\end{figure}

To evaluate the impact of class-prior on the diversity of generated samples, we train the generator~$\theta_G$ using a trained ResNet-34 model on CIFAR-10, while varying the intensity parameter of $\L_{Cls}$~(i.e., $\alpha$ in Eq. (\ref{eqn:lg:prev})) and fixing those of $\L_{Rep}$ and $\L_{Adv}$. To measure the sample diversity, we compute the \textit{Frechet Inception distance} (FID) score over the samples generated by each trained version of $\theta_G$, where the FID score~\cite{FID} is known to be smaller when evaluating more diverse and realistic samples in generative models. Figure~\ref{fig:FID_official} shows a roughly decreasing trend of the FID score when reducing the class-prior's intensity, implying that the stronger the class-prior loss, the lower the diversity of generated samples. In Figure~\ref{fig: mode collapse}, we also visually demonstrate that a generator trained with class-prior produces a limited variety of images for \textit{Airplane} and \textit{Dog} classes, whereas it becomes able to generate variants of those images when removing class-prior from the generator loss function. Finally, as shown in Table~\ref{tab:FID_arbitrary}, this trend turns out to remain the same even when using various teacher models of similar performance. With the exception of T4, where class-prior appears to be effective, the FID scores without class-prior are mostly smaller (and thus exhibit higher diversity) than those with class-prior. 

\subsubsection{Incomplete Quality Control.}

\begin{table}[t]
    \centering
    \small
    \begin{tabular}{c|ccccc}
    \toprule
    \multirow{2}{*}{Teachers} & T1 & T2  & T3  & T4  & T5  \\
     & (95.5) & (93.5) & (92.0) & (94.4) & (91.4) \\
    \midrule
    $\alpha=0$ & \textbf{116.3} & \textbf{119.3} & \textbf{106.7} & 109.7 & \textbf{106.4} \\
    $\alpha=1$ & 122.7 & 122.9 & 111.7 & \textbf{105.1} & 109.6 \\
    \bottomrule             
    \end{tabular}
    \caption{FID scores using five different ResNet-34 teacher models on CIFAR-10.}
    \label{tab:FID_arbitrary}
\end{table}

\begin{figure}[t!]
    \centering
    \includegraphics[width=0.95\columnwidth]{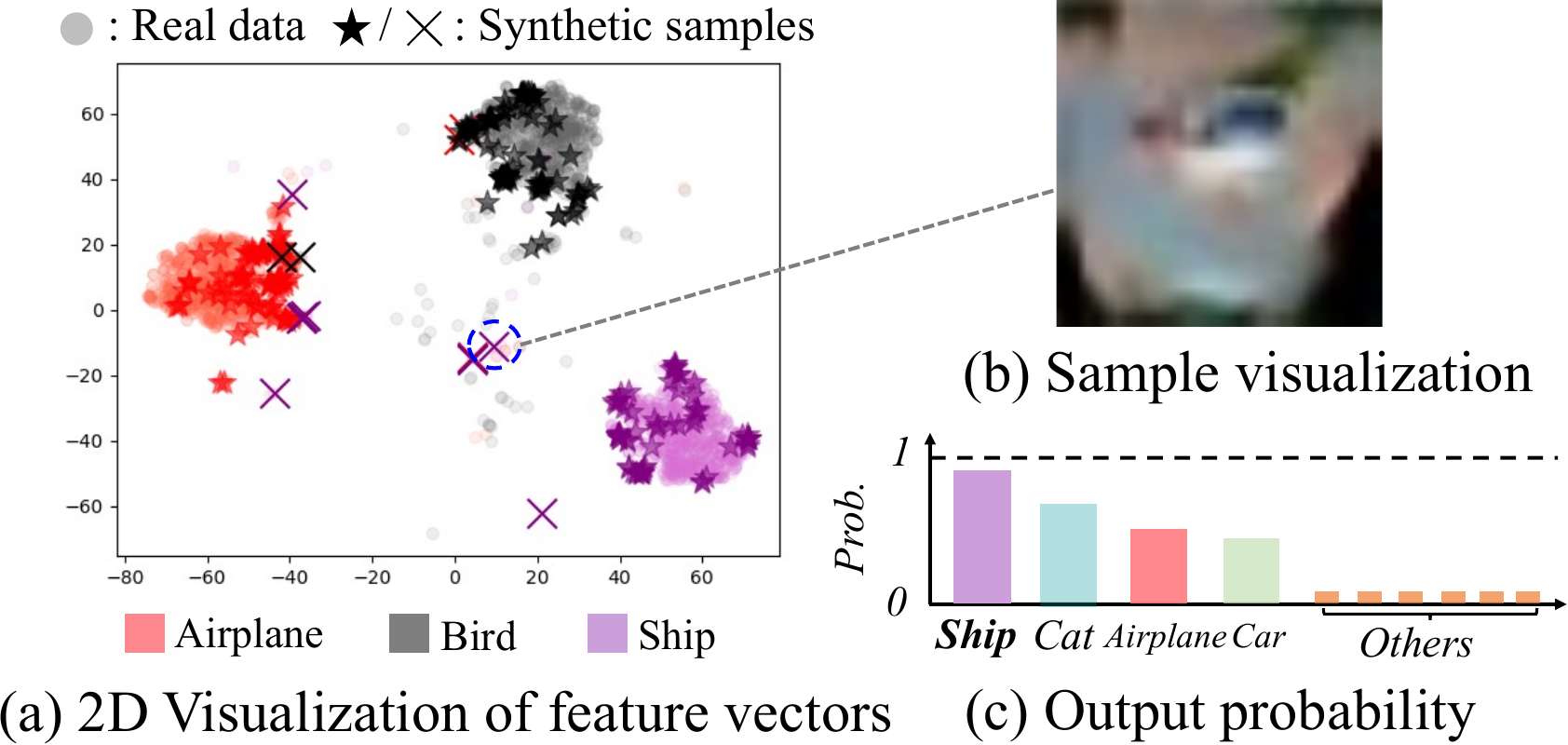}
    \caption{(a) 2D visualization of feature vectors corresponding to real data and synthetic data generated by a generator trained using class-prior without the adversarial loss in ResNet-34 on CIFAR-10, where $\bullet$, $\star$, and $\times$ represent real data samples, high-quality synthetic samples within the boundary of their corresponding real data, and low-quality ones out of their boundary. (b) and (c) show a low-quality synthetic image and its probability distribution, respectively.}
    \label{fig:Unexpected_sample}
\end{figure}

\begin{figure*}[t]
    \centering
    \hspace{-5mm}\subfigure[\label{fig:overview_a} Overall process of TA-DFKD]{\hspace{5mm}\includegraphics[height=46mm]{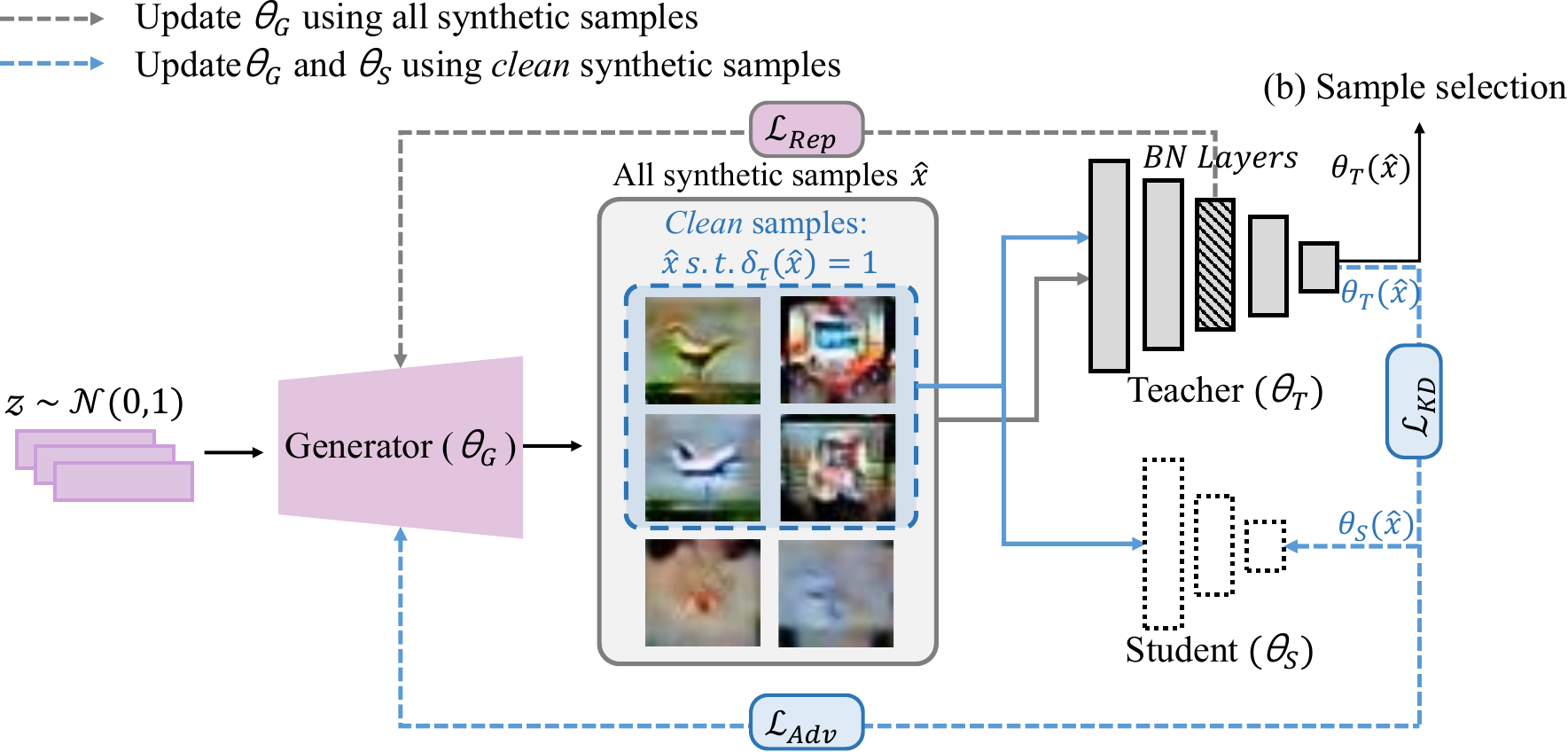}\hspace{5mm}}
    \subfigure[\label{fig:GMM}Sample selection by GMM]{\hspace{5mm}\includegraphics[height=49mm]{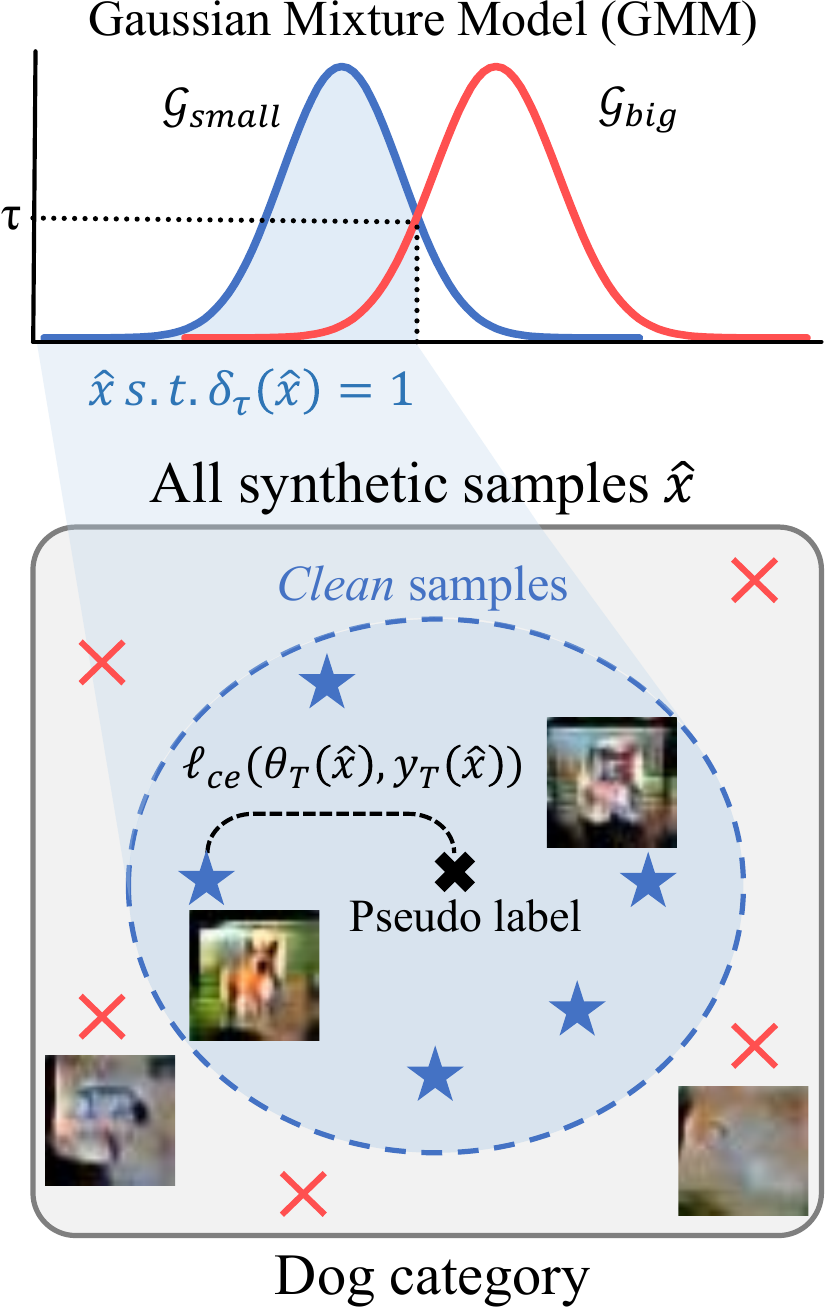}\hspace{5mm}}
    \caption{Overview of the proposed TA-DFKD method.}
    \label{fig:Over_view}
\end{figure*}

We next investigate how effectively the class-prior loss controls the quality of generated samples, by training a generator with class-prior but removing the adversarial loss $\L_{Adv}$ from Eq.~(\ref{eqn:lg:prev}). This is because, as pointed out by~\citet{CMI21}, adversarial training seems to be the major component that causes low-quality samples in DFKD, while class-prior is supposed to enhance the sample quality. As observed in Figure~\ref{fig:Unexpected_sample}, however, even such a generator without $\L_{Adv}$ often synthesizes unexpectedly low-quality samples (represented as $\times$-shaped points in Figure~\ref{fig:Unexpected_sample}(a) and visualized in Figure~\ref{fig:Unexpected_sample}(b)) to the point that the teacher model cannot be confident about their predicted classes (see Figure~\ref{fig:Unexpected_sample}(c)). Unfortunately, these erroneous samples have been consistently observed to account for approximately 7-8\% per batch, and can confuse even well-trained teacher models, potentially leading to the failure of the entire distillation process due to their accumulated errors. Therefore, we conclude that the class-prior loss cannot solely prevent the generation of unexpectedly low-quality samples.

\subsection{Proposed Method}

\subsubsection{Generator Loss Without Class-Prior.}

Based on the limitations of class-prior, our first remedy is to remove $\L_{Cls}$ from $\mathcal{L}_G$, and therefore we have:
\begin{equation} \label{eqn:lg:ours}
\L_G = \beta\L_{Adv} + \gamma\L_{Rep}.
\end{equation}
For $\L_{Adv}$ and $\L_{Rep}$, we first define their individual loss functions, $\ell_{adv}(\hat{x})$ and $\ell_{rep}(\hat{x})$, respectively, for a synthetic sample~$\hat{x}$. Then, $\L_{Adv}$ and $\L_{Rep}$ are simply the expectations of their individual losses over the generated samples.

For the adversarial loss~$\ell_{adv}(\hat{x})$, in common with the recent DFKD methods~\cite{Dreming_to_distill,PREDFKD22,PRE_META}, we use the Jensen-Shannon (JS) divergence~$JSD(\cdot, \cdot)$ as follows:
\begin{equation}
    \ell_{adv}(\hat{x}) = 1 - JSD(\theta_T(\hat{x}), \theta_S(\hat{x})). \nonumber
\end{equation}
Minimizing $\ell_{adv}(\hat{x})$ maximizes the discrepancy between the outputs of the teacher and student models with $\hat{x}$, thereby guiding the generator to produce more difficult samples.

To specify $\ell_{rep}(\hat{x})$, we adopt the BNS technique~\cite{Dreming_to_distill}, which matches the statistics of batch normalization (BN) layers to make generated samples more realistic, by the following definition:
\begin{eqnarray}
    \ell_{rep}(\hat{x}) = \ell_{bns}(\hat{x}) + \lambda \ell_{var}(\hat{x}) + (1-\lambda) \ell_{l2}(\hat{x}), \nonumber
\end{eqnarray}
where $\ell_{bns}(\hat{x})$ is the sum of differences between the statistics stored in BN layers of the teacher model when training real data, $\mu_l$ and $\sigma_l^2$, and those obtained by generated samples in the teacher's same layers, $\mu_l(\hat{x})$ and $\sigma_l^2(\hat{x})$, as: $\ell_{bns}(\hat{x}) = \sum_{l}(\parallel \mu_l(\hat{x}) - \mu_l \parallel_2 + \parallel \sigma_l^2(\hat{x}) - \sigma_l^2 \parallel_2)$. As in the original BNS technique \cite{Dreming_to_distill}, we also leverage additional regularization terms $\ell_{var}$ and $\ell_{l2}$, which are about total variance on pixel values within each image $\hat{x}$ and L2-norm of $\hat{x}$, respectively.

By minimizing both $\L_{Adv}$ and $\L_{Rep}$, the generator can effectively synthesize samples as diverse as possible and mimic the feature-level summary of real data distribution by matching BN statistics.

\subsubsection{Quality Control via Sample Selection.}
Eliminating the class-prior restriction could potentially lead to an even higher risk of generating unexpectedly low-quality samples. Furthermore, the adversarial loss itself has its own problems that need to be addressed, such as the drastic change in the distribution of generated samples, as highlighted by the recent studies~\cite{PREDFKD22,PRE_META,MAD}. To address both issues, we propose a simple yet effective approach: teacher-driven sample selection, which takes only \textit{clean} samples that are confidently verified by the given teacher model. By doing so, from any teacher models, we not only avoid distillation with erroneous samples, but also possibly mitigate drastic changes in the sample distribution.

More specifically, for a generated sample~$\hat{x}$, we measure the quality of $\hat{x}$ by quantifying how confident the teacher model is about its predicted label, denoted by~$y_T(\hat{x})$. To this end, we compute the cross-entropy loss between the teacher's output probability and its one-hot vector of the predicted label as: $\ell_{ce}(\theta_T(\hat{x}),~y_T(\hat{x}))$. This per-sample loss value is then used to determine whether $\hat{x}$ is reliable enough in terms of its label distribution. Instead of making a decision by some absolute comparison, we specifically employ the Gaussian Mixture Model (GMM), inspired by a method of learning with noisy labels~\cite{DIVIDEMIX}. As illustrated in Figure~\ref{fig:GMM}, for each sample batch, the GMM is built upon per-sample loss values, thereby forming two Gaussian distribution components, namely $\G_{small}$ and $\G_{big}$. The $\G_{small}$ component corresponds to the samples with smaller loss values, which thus are considered to be high-quality samples, while the samples belonging to $\G_{big}$ are likely to be low-quality ones. To determine whether to select $\hat{x}$ or not, we compute its posterior probability~$Pr(\G_{small}|\ell_{ce}(\theta_T(\hat{x}),~y_T(\hat{x})))$ and check if the probability exceeds a specified threshold~$\tau$. This enables us to define the following Boolean function~$\delta_{\tau}(\hat{x})$:
\begin{equation}
\delta_{\tau}(\hat{x})=\left\{
\begin{matrix*}[l]
1 & \text{if}~Pr(\G_{small}|\ell_{ce}(\theta_T(\hat{x}),~y_T(\hat{x}))) > \tau,\\
0 & \text{otherwise}.
\end{matrix*}\right. \nonumber
\end{equation}

\begin{figure}[t!]
    \centering
    \includegraphics[width=0.8\columnwidth]{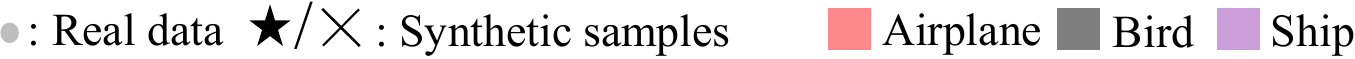}
    
    \subfigure[\label{fig:cp0} Before sample selection ]{\includegraphics[width=0.45\columnwidth]{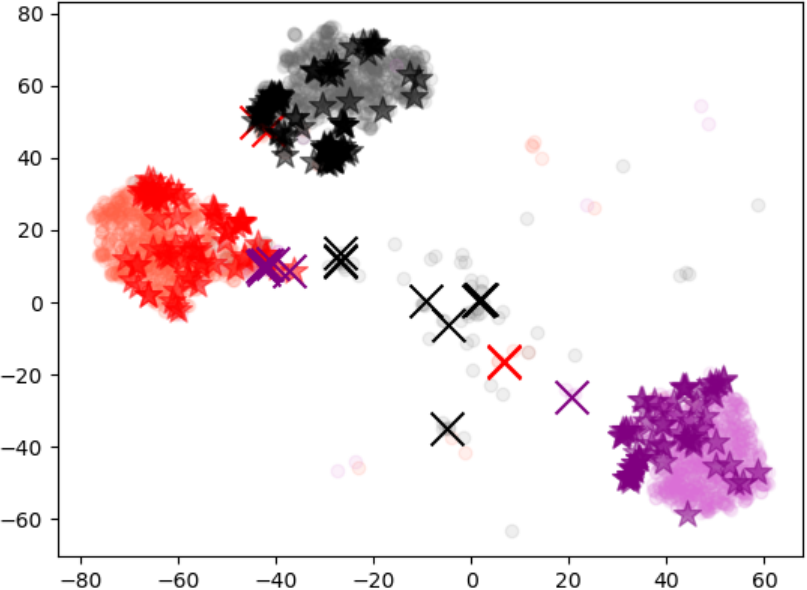}}
    \subfigure[\label{fig:cp0_ss} After sample selection]{\includegraphics[width=0.45\columnwidth]{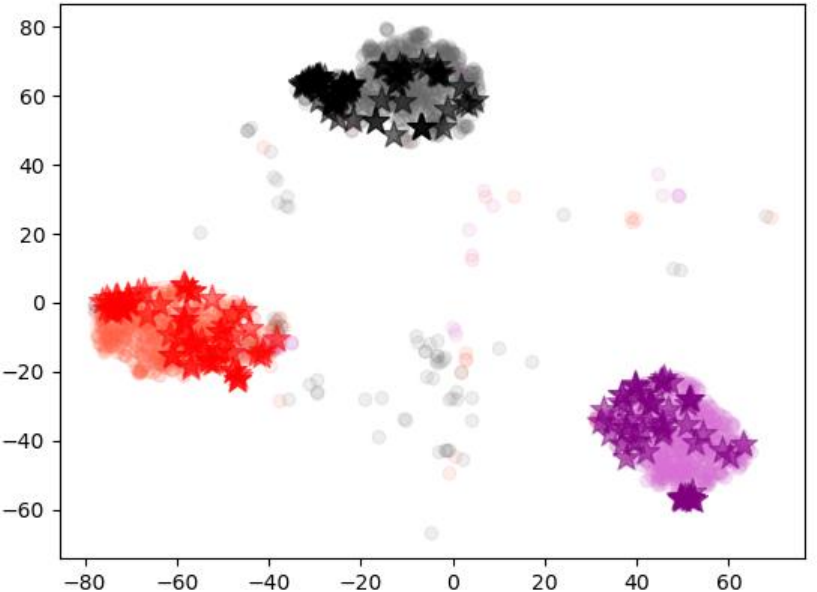}}
    \caption{ Visualization of before and after sample selection from real validation data and synthetic samples in the teacher model ResNet-34 on CIFAR-10.}

    \label{fig:before_anfter_selection }

\end{figure}

Finally, given a set of generated samples, we select only the subset of samples with $\delta_{\tau}(\hat{x}) = 1$ as: 
$$\{\hat{x}~|~\hat{x} = \theta_G(z)~\text{s.t.}~~z \sim p_z(z)~\wedge~\delta_{\tau}(\hat{x}) = 1 \}.$$
Figure~\ref{fig:before_anfter_selection } demonstrates the effectiveness of our sample selection in DFKD. Before applying sample selection to the generator being trained with Eq.~(\ref{eqn:lg:ours}), we can still observe unexpectedly low-quality synthetic samples, as indicated by $\times$-shaped points in Figure~\ref{fig:cp0}. However, they are effectively removed from the result of Figure~\ref{fig:cp0_ss} after our sample selection method is applied, and therein all the synthetic samples are properly located within their corresponding boundary of real data samples. In our experiments, setting $\tau$ to 0.5, in the early stages of training, approximately 60\% of samples are selected, but when approaching the end of training, more than 90\% of samples are selected.

\subsubsection{Overall Process of TA-DFKD.}
We now present the overall process of our teacher-agnostic data-free knowledge distillation (TA-DFKD) method, as illustrated in Figure~\ref{fig:overview_a}. When training the generator with our loss function in Eq.~(\ref{eqn:lg:ours}), we use all the synthetic samples without sample selection to compute $\L_{Rep}$, as $\L_{Rep}$ is for learning feature-level summary of real data distribution. On the other hand, in terms of both $\L_{Adv}$ and $\L_{KD}$, we train with only selected samples by our selection method. Therefore, we accordingly define the following two loss functions of Eq.~(\ref{eqn:lg:ours}):
\begin{eqnarray}
    \L_{Adv} & = & \mathbb{E}_{z \sim p_z(z)~\wedge~\delta_{\tau}(\theta_G(z)) = 1}[\ell_{adv}(\theta_G(z))],~~\text{and}\nonumber \\
    \L_{Rep} & = & \mathbb{E}_{z \sim p_z(z)}[\ell_{rep}(\theta_G(z))]. \nonumber
\end{eqnarray}
The final KD loss is similarly defined as:
$$
\L_{KD} = \mathbb{E}_{z \sim p_z(z)~\wedge~\delta_{\tau}(\theta_G(z)) = 1}\parallel \theta_T(\theta_G(z)) - \theta_S(\theta_G(z)) \parallel_1,
$$
where we use the L1-distance between two outputs using only selected synthetic samples.

\section{Experiments}
In this section, we validate our TA-DFKD method, with a focus on its robustness and stability in DFKD using various pretrained teacher models with the similar test performance.

\subsection{Environment}

\subsubsection{Datasets and Compared Methods.}
We use three benchmark datasets, CIFAR-10/CIFAR-100 \cite{cifar_datasets} and TinyImageNet \cite{imagenet_datasets}. The CIFAR datasets contain 60,000 RGB images of $32 \times 32$ over either 10 or 100 classes, whereas Tiny-ImageNet consists of 100,000 images for 200 classes, 500 for each class, all of which is the same size of $64 \times 64$. Using these datasets, we compare TA-DFKD with multiple SOTA DFKD methods, which are two fold. The first category includes DAFL~\cite{DAFL019} and DFAD~\cite{DFAD}, which partially mix out of the three loss terms, class-prior, adversarial, and representation losses. For the second category, we test CMI~\cite{CMI21} as a representative one using the BNS loss, and PRE-DFKD~\cite{PREDFKD22} that alternatively uses activation maximization for the same purpose. Except for CMI~\cite{CMI21}, which requires an additional training phase with pre-generated samples in memory, all the compared methods are one-phase DFKD methods. Due to the space limit, the results using TinyImageNet are presented in the Appendix.

\subsubsection{Training Details.}
For all datasets, we train ResNet-34~\cite{resnet} as the teacher model, ResNet-18 as the student model, and DCGAN~\cite{DCGAN} as the generator. During training ResNet-34, multiple teacher models with similar performance are randomly selected. In the entire DFKD process, we train ResNet-18 along with DCGAN for a particular number of epochs, 200 epochs for CIFAR-10 and 500 epochs for CIFAR-100 and Tiny-ImageNet. For compared methods, we follow the same configuration of their implementations. Every measurement in this section is taken out of 4 repeated runs.

\subsubsection{Evaluation Metrics.}
In order to evaluate the robustness and stability of each method, we not only measure the peak accuracy~$acc_{max}$ of each student model over all the repeated runs but also introduce the \textit{converging} accuracy~$acc_{last[k]}$, which is the average student accuracy over the last $k$ epochs of KD training for each run. Small differences between $acc_{max}$ and $acc_{last[k]}$ imply that the student model shows stable and reasonably good performance during the last phase of training. Furthermore, a small deviation of $acc_{last[k]}$ out of all the repeated runs indicates a high level of the robustness within a particular teacher model. We set $k$ to 10 for CIFAR-10 and 20 for the other datasets.

\begin{table*}[t]
\begin{center}
    \small
    \resizebox{\textwidth}{!}{
    \begin{tabular}{l|cc|cc|cc|cc|cc}
        \toprule
        \multicolumn{11}{c}{CIFAR-10~~~~~~Teacher: ResNet-34 ~~  Student: ResNet-18 (Accuracy with real data: 95.2 \%)} \\
        \midrule
        \multirow{2}{*}{Method} & \multicolumn{2}{c|}{Teacher 1 (95.46\%)} & \multicolumn{2}{c|}{Teacher 2 (93.46\%)} & \multicolumn{2}{c|}{Teacher 3 (92.01\%)} & \multicolumn{2}{c|}{Teacher 4 (94.43\%)} & \multicolumn{2}{c}{Teacher 5 (91.36\%)} \\
        
        & $acc_{last[10]}$ & $acc_{max}$ & $acc_{last[10]}$ & $acc_{max}$ & $acc_{last[10]}$ & $acc_{max}$ & $acc_{last[10]}$ & $acc_{max}$ & $acc_{last[10]}$ & $acc_{max}$ \\
        
        \midrule
        DAFL & 83.60$_{\pm 7.8}$ & 92.07 & 85.94$_{\pm 2.0}$ & 88.43 & 68.87$_{\pm 20.0}$ & 88.08 & 89.21$_{\pm 4.6}$ & 92.90 & 72.85$_{\pm 10.9}$ & 85.59 \\
        
        DFAD & \underline{93.23}$_{\pm 0.1}$ & 93.60 & 87.72$_{\pm 0.2}$ & 88.69 & 87.83$_{\pm 0.1}$ & 88.77& 92.27$_{\pm 0.1}$ & 92.82 & 87.66$_{\pm0.2}$ & 89.22 \\
        
        CMI & 92.54$_{\pm 1.6}$ & \textbf{94.80} & \underline{89.84}$_{\pm 0.1}$ & \underline{90.16} & \underline{89.40}$_{\pm 0.1}$ & \underline{89.81} & \underline{93.26}$_{\pm 0.1}$ & \underline{93.61} & \underline{89.62}$_{\pm 0.1}$ & \underline{90.37} \\
        
        PRE-DFKD & 89.22$_{\pm 4.9}$ & 94.10 & 85.16$_{\pm 0.3}$ & 86.59 & 80.55$_{\pm 0.6}$ & 83.15 & 90.80$_{\pm 0.3}$ & 91.56 & 83.93$_{\pm 1.6}$ & 88.05 \\
        
        TA-DFKD & \textbf{94.24$_{\pm 0.1}$} & \underline{94.43} & \textbf{91.99$_{\pm 0.1}$} & \textbf{92.15} & \textbf{90.21$_{\pm 0.1}$} & \textbf{90.69} & \textbf{93.61$_{\pm 0.0}$} & \textbf{93.79} & \textbf{90.27$_{\pm 0.1}$} & \textbf{91.08} \\
        \bottomrule
    


        \toprule
        \multicolumn{11}{c}{CIFAR-100~~~~~~Teacher: ResNet-34 ~~  Student: ResNet-18 (Accuracy with real data: 77.1 \%)} \\
        \midrule
        \multirow{2}{*}{Method} & \multicolumn{2}{c|}{Teacher 1 (77.98\%)} & \multicolumn{2}{c|}{Teacher 2 (75.01\%)} & \multicolumn{2}{c|}{Teacher 3 (76.01\%)} & \multicolumn{2}{c|}{Teacher 4 (78.42\%)} & \multicolumn{2}{c}{Teacher 5 (77.04\%)} \\
        
        & $acc_{last[20]}$ & $acc_{max}$ & $acc_{last[20]}$ & $acc_{max}$ & $acc_{last[20]}$ & $acc_{max}$ & $acc_{last[20]}$ & $acc_{max}$ & $acc_{last[20]}$ & $acc_{max}$ \\
        
        \midrule
        DAFL & 74.08$_{\pm 0.6}$ & 75.22 & 70.31$_{\pm 0.4}$ & 71.27 & 72.22$_{\pm 1.0}$ & 73.73 & 73.96$_{\pm 0.5}$ & 74.82 & 74.66$_{\pm 0.4}$ & 75.32 \\
        
        DFAD & 69.51$_{\pm 0.3}$ & 70.03 & 66.31$_{\pm 0.1}$ & 66.75 & 71.55$_{\pm 0.2}$ & 71.97 & 69.61$_{\pm 0.3}$ & 70.26 & 70.33$_{\pm 0.4}$ & 71.33 \\
        
        CMI & 74.10$_{\pm 0.2}$ & 74.85 & 71.81$_{\pm 0.1}$ & 72.43 & 73.55$_{\pm 0.1}$ & 74.03 & 74.40$_{\pm 0.1}$ & \underline{77.00} & 74.18$_{\pm 0.1}$ & 74.77 \\
        
        PRE-DFKD & \underline{76.13}$_{\pm 0.2}$ & \underline{76.57} & \underline{73.11}$_{\pm 0.2}$ & \underline{73.53} & \underline{74.75}$_{\pm 0.4}$ & \underline{75.44} & \underline{75.58}$_{\pm 1.1}$ & \textbf{77.10} & \underline{75.37}$_{\pm 0.6}$ & \underline{76.01} \\
        
        TA-DFKD & \textbf{76.55$_{\pm 0.1}$} & \textbf{76.76} & \textbf{73.61$_{\pm 0.1}$} & \textbf{73.89} & \textbf{75.74$_{\pm 0.1}$} & \textbf{76.02} & \textbf{76.73$_{\pm 0.1}$} & 76.99 & \textbf{76.58$_{\pm 0.1}$} & \textbf{76.84} \\
        \bottomrule
    \end{tabular}}

\caption{DFKD performance comparison using 5 teacher models trained on CIFAR-10 (top) and  CIFAR-100 (bottom).}
\label{tab:Results of CIFAR}
\end{center}  
\end{table*}

\subsection{Experimental Results}

\begin{figure}[t!]
    \centering
    \includegraphics[width=0.6\columnwidth]{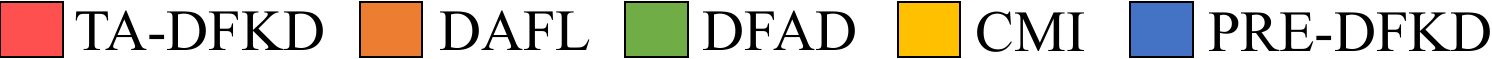}
    \subfigure[\label{fig:CIFAR10_Robustness} CIFAR-10]{\includegraphics[height=27mm]{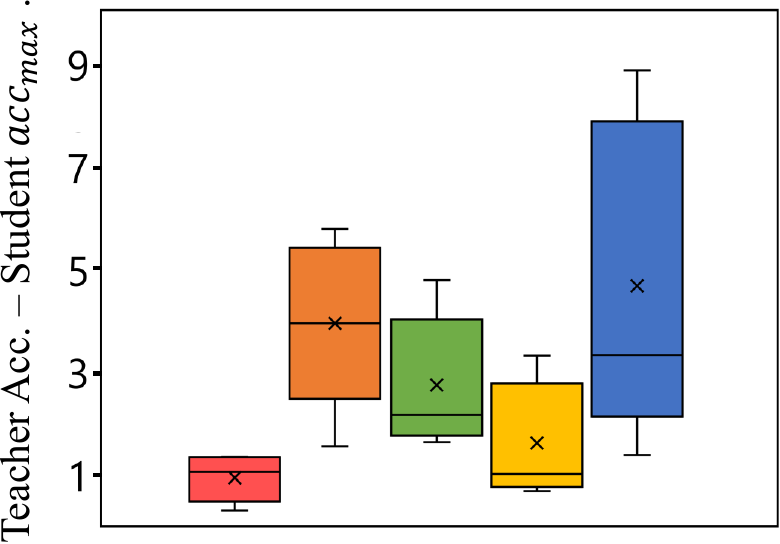}}
    \hspace{-32mm}
    \hfill\subfigure[\label{fig:CIFAR100_Robustness} CIFAR-100]{\includegraphics[height=27mm]{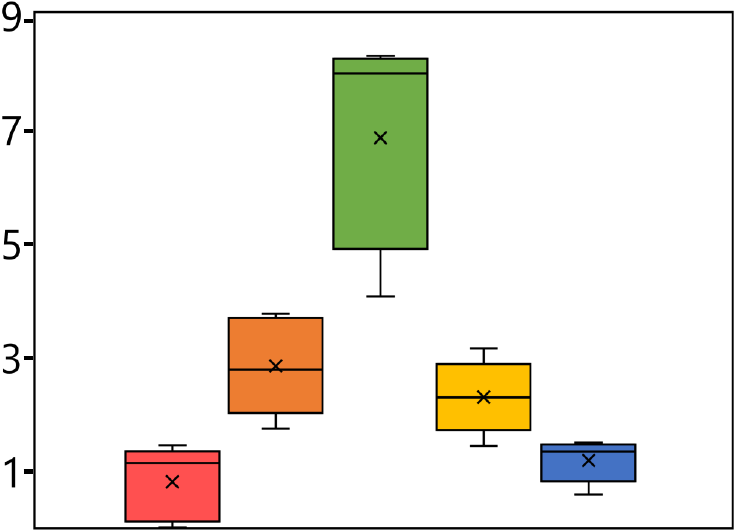}}
    \caption{Box-plots of performance differences between teacher and student in the CIFAR datasets.}
    \label{fig:Robustness}

    \includegraphics[width=0.6\columnwidth]{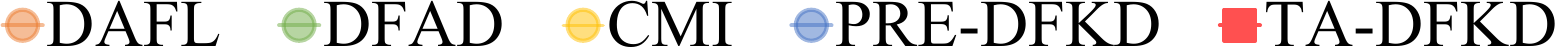}
    \subfigure[\label{fig:CIFAR10_Evaluation metrics} CIFAR-10]{\includegraphics[height=24mm]{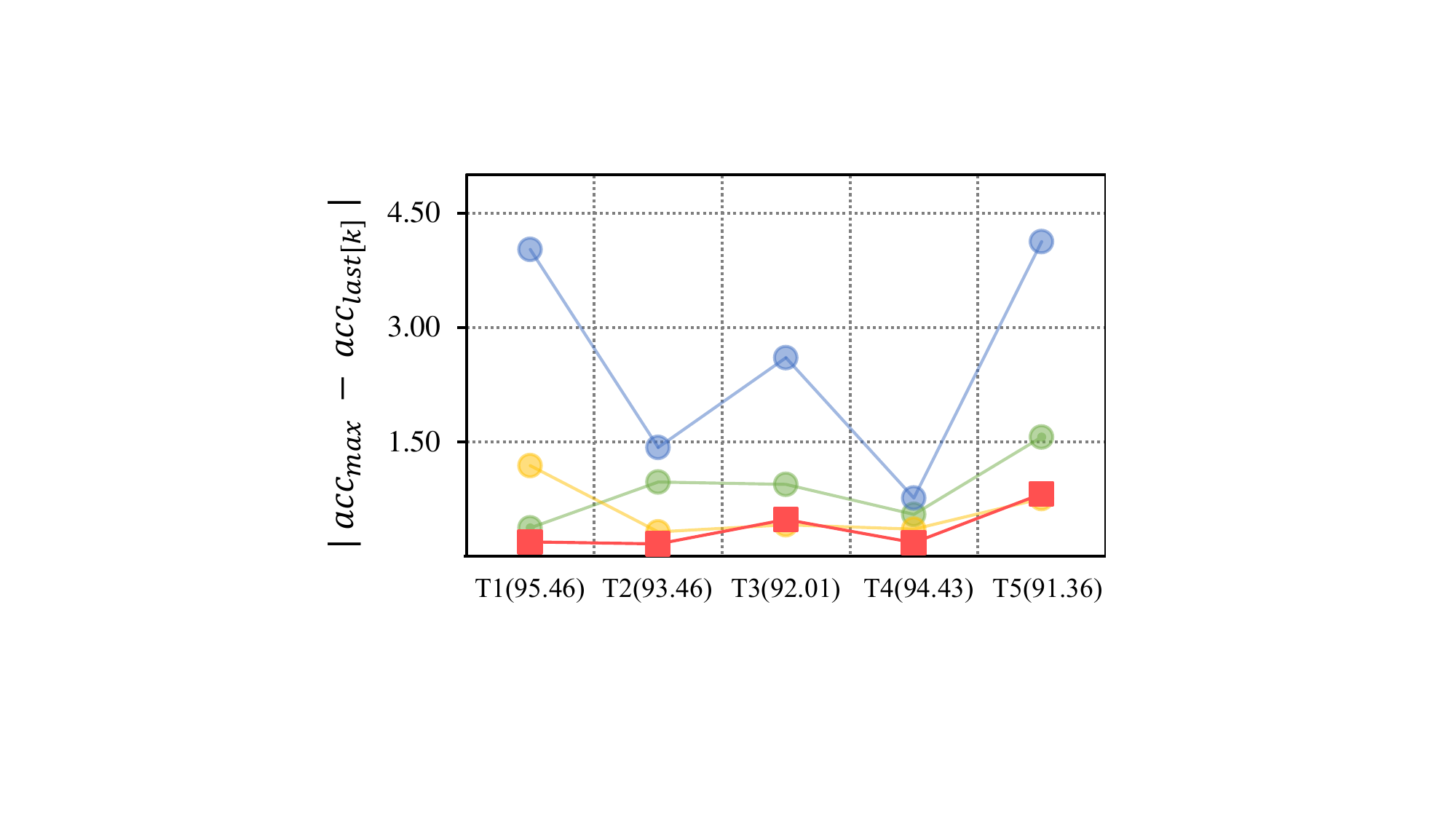}}
    \hfill
    \subfigure[\label{fig:CIFAR100_Evaluation metrics} CIFAR-100]{\includegraphics[height=24mm]{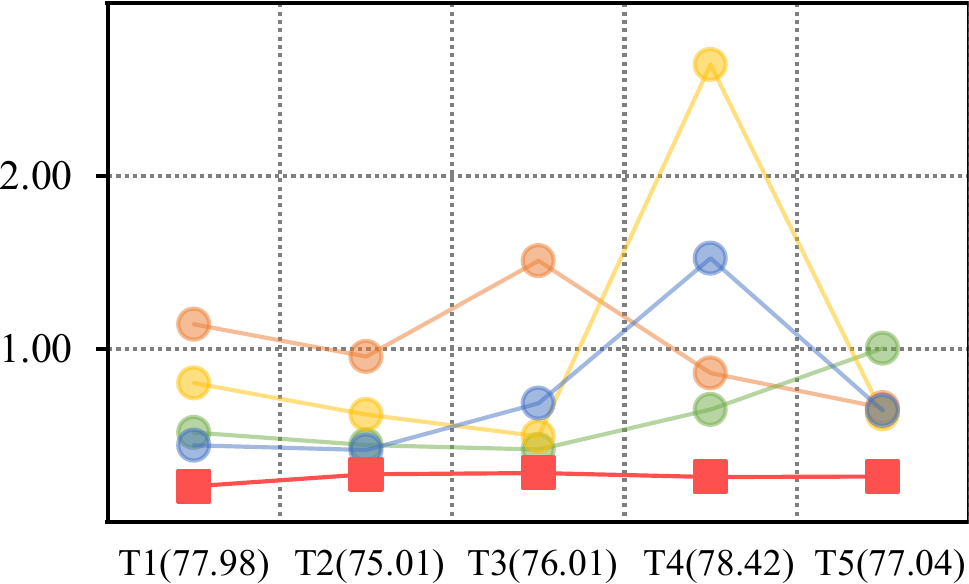}}
    \caption{Differences between $acc_{max}$ and $acc_{last[k]}$ in the CIFAR datasets.}
    \label{fig:Evaluation metrics}
\end{figure}

\subsubsection{Performance Comparison.}
Table~\ref{tab:Results of CIFAR} presents the summarized result of performance comparison of TA-DFKD with the SOTA DFKD methods, using five different teacher models trained on the CIFAR-10 and CIFAR-100 datasets. It is clearly observed that TA-DFKD manages to achieve the highest peak accuracy~$acc_{max}$ as well as the highest converging accuracy~$acc_{last[k]}$ in most of the cases. Over all the repeated runs, TA-DFKD shows only small variations in its converging accuracy, implying high robustness within each teacher model. On the other hand, DAFL sometimes experiences a catastrophic failure of distillation with a large deviation even with the same teacher model (e.g., $\pm~20.02$ in Teacher 3 on CIFAR-10), aligning with the example of Figure \ref{fig:problem_trim}(a). This failure is likely to happen when the generator gets collapsed into only a few easy samples at some point of training. Recent methods utilizing all the three loss terms, CMI and PRE-DFKD, generally perform better than those of not using all the terms, DAFL and DFAD. However, both CMI and PRE-DFKD do not show the reliable performance across the two datasets in that either of them interchangeably takes the second best position in different datasets.

\subsubsection{Teacher-Agnostic Behavior.}
Based on the results of Table~\ref{tab:Results of CIFAR}, we examine how robust and stable the performance of each method remains when using different teacher models, as plotted in Figures \ref{fig:Robustness} and \ref{fig:Evaluation metrics}. Figure~\ref{fig:Robustness} shows box-plots on performance gaps in peak accuracies between the teacher and student models. In both CIFAR-10 and CIFAR-100, the proposed TA-DFKD shows short box-plots implying the teacher-agnostic robustness, while the other compared methods have relatively long ranges of performance gaps throughout different teacher models. Figure~\ref{fig:Evaluation metrics} demonstrates the teacher-agnostic stability by plotting differences between $acc_{max}$ and $acc_{last[k]}$ using five teacher models. TA-DFKD clearly takes the bottom-most position in both graphs of the CIFAR datasets, meaning that its performance becomes quite stable and remains almost the same as its best accuracy once it reaches the last phase of distillation process.

\begin{table}[t]
    \centering
    \small
    \resizebox{0.46\textwidth}{!}{
    \begin{tabular}{c|ccccc}

    \toprule
    
    \multirow{2}{*}{Method} & T1 & T2 & T3 & T4 & T5 \\ 
    & (77.98) & (75.01) & (76.01) & (78.42) & (77.04) \\

    \midrule

    Baseline & 73.67 & 71.09 & 74.22 & 73.14 & 76.08 \\
    w/o $\L_{Cls}$& \underline{73.96} & \underline{71.23}& \underline{74.43} & \underline{76.08} & \underline{76.67} \\
    TA-DFKD & \textbf{76.76} & \textbf{73.89} & \textbf{76.02} & \textbf{76.99} & \textbf{76.84}\\

    \bottomrule
    \end{tabular}}
    \caption{Ablation study showing peak accuracies on CIFAR-100, where (1) baseline is the standard DFKD framework involving all the three loss terms, (2) w/o class-prior is the method removing class-prior from the standard framework, and (3) TA-DFKD is our final version additionally applying teacher-driven sample selection.}
    \label{tab: ablation_study}
\end{table}

\subsubsection{Ablation Study.}
Table~\ref{tab: ablation_study} shows the result of an ablation study to verify the effectiveness of elimination of class-prior and applying sample selection, using the five teacher models on CIFAR-100. The baseline methods use all the three loss terms, $\mathcal{L}_{Cls}$, $\mathcal{L}_{Rep}$, and $\mathcal{L}_{Adv}$, without any sample selection. The result clearly confirms our two arguments: (1) class-prior is better to be removed, but (2) removing class-prior is not sufficient to further improve the performance without controlling sample quality by our sample selection method.

\section{Conclusion}
This paper has conducted the first study on teacher-agnostic DFKD, with a focus on three loss terms commonly adopted in DFKD methodologies. Our findings strongly suggest that by replacing the class-prior restriction with our sample selection scheme, we can achieve enhanced quality control, thus leading us to propose the TA-DFKD method. In our experiments, TA-DFKD has demonstrated remarkable robustness and stability across various teacher models. We believe that our work offers a practical solution for knowledge distillation scenarios without access to prior data samples, and it is our hope that this work marks the initiation of the problem of teacher-agnostic DFKD, providing a promising direction for further research in the field.

\section{Acknowledgments}
This work was supported in part by Institute of Information \& communications Technology Planning \& Evaluation (IITP) grants funded by the Korea government(MSIT) (No.2022-0-00448, Deep Total Recall: Continual Learning for Human-Like Recall of Artificial Neural Networks, No.RS-2022-00155915, Artificial Intelligence Convergence Innovation Human Resources Development (Inha University)), in part by the National Research Foundation of Korea (NRF) grants funded by the Korea government (MSIT) (No.2021R1F1A1060160, No.2022R1A4A3029480), and in part by INHA UNIVERSITY Research Grant.

\bibliography{aaai24}


\onecolumn
\setcounter{table}{0}
\setcounter{figure}{0}
\renewcommand{\thepage}{S\arabic{page}} 
\renewcommand{\thesection}{S\arabic{section}}  
\renewcommand{\thetable}{A\arabic{table}}  
\renewcommand{\thefigure}{A\arabic{figure}}

\appendix
\section{Appendix of  ``Teacher as a Lenient Expert: Teacher-Agnostic Data-Free Knowledge Distillation''}

In this appendix, we first (1) present additional experimental performance comparison using 2 teacher models trained on Tiny-ImageNet, and (2) more detailed experimental results on the ablation study, and then (3) provide detailed values of Figures~\ref{fig:Robustness}~and~\ref{fig:Evaluation metrics}. Finally, we (4) describe all the implementation details and hyperparameter values.

\section{Details of Experimental Results}

\subsection{Results of Tiny-ImageNet.}
Table~\ref{tab: result of tinyimagenet} shows the performance summary of the experiments using Tiny-ImageNet. As mentioned in \cite{PREDFKD22}, we have also failed to find proper hyperparameters for CMI and DFAD, and therefore we compare TA-DFKD with only DAFL and PRE-DFKD. Similar to the results of the CIFAR datasets, our TA-DFKD method outperforms the compared methods in both the robustness and stability across two teacher models. Notably, PRE-DFKD exhibits significant performance variations in the same teacher model (e.g., $\pm~9.78$ in Teacher~1). This is another evidence that using both class-prior and adversarial losses may not always reach to a proper balance between diversity and sample quality.

\begin{table}[h]
    \centering
    \scriptsize
    \begin{tabular}{l|cc|cc}
    \toprule

    \multicolumn{5}{c}{Tiny-ImageNet}\\
    \multicolumn{5}{c}{Teacher: ResNet-34 ~~ Student: ResNet-18 (Accuracy with real data: 64.9 \% )}\\
    \midrule
    \multirow{2}{*}{Method} & \multicolumn{2}{c|}{Teacher 1 (75.50\%)} & \multicolumn{2}{c}{Teacher 2 (74.92\%)} \\
    & $acc_{last[20]}$ & $acc_{max}$ & $acc_{last[20]}$ & $acc_{max}$\\
    \midrule
    DAFL & \underline{47.76}$_{\pm2.06}$ & 51.33 & \underline{50.60}$_{\pm 1.84}$ & 53.20 \\
    
    PRE-DFKD & 46.45$_{\pm 9.78}$ & \underline{53.16} & 46.13$_{\pm 7.50}$ & \underline{53.68} \\
    
    TA-DFKD (ours) & \textbf{53.00$_{\pm 1.57}$} & \textbf{54.84} & \textbf{53.55$_{\pm 0.49}$} & \textbf{54.52} \\
    \bottomrule

    \end{tabular}
    \caption{Results of Tiny-ImageNet in two teacher models}
    \label{tab: result of tinyimagenet}
    
\end{table}

\subsection{Detailed Ablation Study.}

We conduct an ablation study using the CIFAR-10 dataset and report the averaged outcomes from 4 repeated runs, rather than focusing only on maximum values. Similar to the results of Table~\ref{tab: ablation_study} using the CIFAR-100 dataset, we can demonstrate the effectiveness of applying sample selection and removing class-prior.

\begin{table}[h]
    \centering
    \scriptsize
    \begin{tabular}{c|ccccc}

    \toprule
    \multirow{2}{*}{Method} & \multicolumn{5}{c}{CIFAR-10} \\
    \cmidrule{2-6}
    
    & Teacher 1 (95.46\%) & Teacher 2 (93.46\%) & Teacher 3 (92.01\%) & Teacher 4 (94.43\%) & Teacher 5 (91.36\%) \\ 

    \midrule

    Baseline & \underline{93.25}$_{\pm 0.14}$ & 90.54$_{\pm 0.18}$ & 88.52$_{\pm 0.17}$ & \underline{92.88}$_{\pm 0.07}$ & 88.90$_{\pm 0.15}$ \\
    w/o class-prior & 93.06$_{\pm 0.09}$ & \underline{90.61}$_{\pm 0.12}$ & \underline{88.71}$_{\pm 0.12}$ & 92.84$_{\pm 0.14}$ & \underline{90.55}$_{\pm 0.10}$ \\
    TA-DFKD (ours) & \textbf{94.35}$_{\pm 0.06}$ & \textbf{92.10}$_{\pm 0.06}$ & \textbf{90.46}$_{\pm 0.15}$ & \textbf{93.72}$_{\pm 0.07}$ & \textbf{90.85}$_{\pm 0.20}$\\

    \bottomrule
    \end{tabular}
    \vspace{2mm}

    \begin{tabular}{c|ccccc}

    \toprule
    \multirow{2}{*}{Method} & \multicolumn{5}{c}{CIFAR-100} \\
    \cmidrule{2-6}
    
    & Teacher 1 (77.98\%) & Teacher 2 (75.01\%) & Teacher 3 (76.01\%) & Teacher 4 (78.42\%) & Teacher 5 (77.04\%) \\ 

    \midrule

    Baseline & 73.67$_{\pm 0.34}$ & 71.09$_{\pm 0.22}$ & 74.22$_{\pm 0.10}$ & 73.14$_{\pm 0.25}$ & 73.64$_{\pm 0.15}$ \\
    w/o class-prior & \underline{73.96}$_{\pm 0.19}$ & \underline{71.23}$_{\pm 0.21}$ & \underline{74.43}$_{\pm 0.16}$ & \underline{76.08}$_{\pm 0.26}$ & \underline{76.67}$_{\pm 0.08}$ \\
    TA-DFKD (ours) & \textbf{76.74}$_{\pm 0.02}$ & \textbf{73.83}$_{\pm 0.06}$ & \textbf{75.93}$_{\pm 0.06}$ & \textbf{76.32}$_{\pm 0.40}$ & \textbf{76.79}$_{\pm 0.03}$\\

    \bottomrule
    \end{tabular}
    \caption{Detailed results of the ablation study using CIFAR-10 and CIFAR-100}
    \label{tab: ablation_study_detail}
   
\end{table}

\subsection{Detailed Values in Figure~\ref{fig:Robustness} and Figure~\ref{fig:Evaluation metrics}}

Table~\ref{tab: fig7_detail} and Table~\ref{tab: fig8_detail} present detailed values corresponding to the graphs shown in Figures~\ref{fig:Robustness} and \ref{fig:Evaluation metrics}, respectively. Table~\ref{tab: fig7_detail} specifically illustrate the differences between the accuracy of the teacher and the peak accuracy of the student denoted as $acc_{max}$.  TA-DFKD achieves the best teacher-agnostic robustness by almost always exhibiting the smallest variance from the corresponding teacher's accuracy. Table~\ref{tab: fig8_detail} shows differences between the peak accuracy ($acc_{max}$) and the converging accuracy ($acc_{last[k]}$) within each student model, in order to assess the stability of the student model during its converging phase of training. In most of the cases, our proposed TA-DFKD method demonstrates superior stability in performance, evidenced by minimal deviations between the best accuracy and the average accuracy over the last epochs.

\begin{table}[h]
    \centering
    \scriptsize
    \begin{tabular}{c|ccccc|ccccc}

    \toprule
    \multirow{2}{*}{Method} & \multicolumn{5}{c|}{CIFAR-10} & \multicolumn{5}{c}{CIFAR-100} \\
    \cmidrule{2-11}
    & T1(95.46\%) & T2(93.46\%) & T3(92.01\%) & T4(94.43\%) & T5(91.36\%) & T1(77.98\%) & T2(75.01\%) & T3(76.01\%) & T4(78.42\%) & T5(77.04\%) \\ 

    \midrule
    DAFL & 3.39 & 5.03 & 3.93  & 1.53 & 5.77 & 2.76 & 3.74 & 2.28 & 3.60 & 1.72 \\
    DFAD & 1.86 & 4.77 & 3.24 & 1.61 & 2.14 & 7.95 & 8.26 & 4.04 & 8.16 & 5.71 \\
    CMI & \textbf{0.66} & \underline{3.30} & \underline{2.20} & \underline{0.82} & \underline{0.99} & 3.13 & 2.58 & 1.98 & \underline{1.42} & 2.27 \\
    PRE-DFKD &1.36 & 6.87 &8.86 & 2.87 & 3.31 & \underline{1.41} & \underline{1.48} & \underline{0.57} & \textbf{1.32} & \underline{1.03} \\  
    TA-DFKD & \underline{1.03} & \textbf{1.31}& \textbf{1.32} & \textbf{0.64} & \textbf{0.28} & \textbf{1.22} & \textbf{1.12} & \textbf{-0.01} & 1.43 & \textbf{0.20} \\
    
    \bottomrule
    
    \end{tabular}
    \caption{Differences in performance between teachers and student models (Teacher Acc. - Student Peak Acc. ($acc_{max}$)).}
    \label{tab: fig7_detail}
    \vspace{4mm}

    \scriptsize
    \begin{tabular}{c|ccccc|ccccc}
    \toprule

    \multirow{2}{*}{Method} & \multicolumn{5}{c|}{CIFAR-10} & \multicolumn{5}{c}{CIFAR-100} \\
    \cmidrule{2-11}
    & T1(95.46\%) & T2(93.46\%) & T3(92.01\%) & T4(94.43\%) & T5(91.36\%) & T1(77.98\%) & T2(75.01\%) & T3(76.01\%) & T4(78.42\%) & T5(77.04\%) \\

    \midrule
    DAFL & 8.47 & 2.49 & 19.21  & 3.69 & 12.74 & 1.14 & 0.96 & 1.51 & 0.86 & 0.66 \\
    DFAD & \underline{0.37} & 0.97 & 0.94 & 0.55 & 1.56 & 0.52 & 0.44 & \underline{0.42} & \underline{0.65} & 1.00 \\
    CMI & 2.26 & \underline{0.32} & \textbf{0.41} & \underline{0.35} & \textbf{0.75} & 0.75 & 0.62 & 0.48 & 2.60 & \underline{0.59} \\
    PRE-DFKD & 4.88 & 1.43 & 2.60 &0.76 & 4.12 & \underline{0.44} & \underline{0.42} & 0.69 & 1.52 &0.64 \\
    TA-DFKD & \textbf{0.19} & \textbf{0.16} & \underline{0.48} &\textbf{0.18} & \underline{0.82} & \textbf{0.21} & \textbf{0.28} & \textbf{0.28} &\textbf{0.26}  & \textbf{0.26} \\
    
    \bottomrule
    
    \end{tabular}
    \caption{Differences between the peak accuracy and converging accuracy within each student model (Peak Acc. ($acc_{max}$) - Converging Acc. ($acc_{last[k]}$).}
    \label{tab: fig8_detail}

\end{table}

\section{Implementation Details}

\subsection{How to Train Various Teacher Models}
Let us provide how we get various teacher models in detail. For the teacher models, except for Tiny-ImageNet, we use the same training environment as used in DAFL~\cite{DAFL019}, referring to its implementation page: \url{https://github.com/autogyro/DAFL/blob/master/teacher-train.py}. Basically, during a relatively longer period of training, we randomly take multiple trained models as long as their performance is reasonably high. 

More specifically, in MNIST, during the training of 100 epochs, we first selected a reference teacher model, so-called Teacher~1 (98.9~\%), whose accuracy is the same as they are in the existing DFKD methods~\cite{DFAD,PREDFKD22}. Then, we have picked up a slightly better version as Teacher~2 (99.2\%). 

In the CIFAR datasets, we increased the number of epochs to be 500 and adjusted the learning rate step decay by the same ratio (reducing by $0.1 \times$ at epochs 200 and 300). As in MNIST, we first selected the models with accuracy 95.46 and 77.46 as the reference teachers (i.e., Teacher~1) for CIFAR-10 and CIFAR-100, respectively, as those accuracies similarly appeared in the existing DFKD methods. For the other teacher models, we randomly selected ones whose accuracies are not below 91\% for CIFAR-10 and 75\% for CIFAR-100. 

For Tiny-ImageNet, we fine-tuned a pretrained model on ImageNet with the Tiny-ImageNet dataset to the point that the resulting accuracy of the model gets reasonably high, such as 75\%, and randomly selected two teacher models.

\subsection{Details of Implementation and Hyperparameters of TA-DFKD}
Our code is available at the following website: \url{https://github.com/bigdata-inha/TA-DFKD-Official}. For fair comparison, we follow the same settings, which include the number of epochs and iterations per epoch for the entire DFKD process, as those of PRE-DFKD~\cite{PREDFKD22}. For all datasets, we use the same set of hyperparameters. More specifically, we set the dimensionality of latent vectors~(i.e., $z$) to 1000, and set the batch size to 1024. When training the generator, we adopted Adam optimizer with $\beta=$~1, $\gamma=$~10 and a learning rate of 0.001. In the distillation of student models, based on SGD optimizer, we used 0.01~initial learning rate with a cosine learning rate decay, 5e-4~weight decay and 0.9~momentum.

\end{document}